\documentclass[sigconf]{acmart}
\usepackage{booktabs}
\usepackage{tabularx}
\usepackage{array}
\setcopyright{none}
\renewcommand\footnotetextcopyrightpermission[1]{}
\acmConference[ICAIF '26 Workshop]{3rd Workshop on LLMs and Generative AI for Finance}{November 14, 2026}{Milan, Italy}
\newcolumntype{L}[1]{>{\hsize=#1\hsize\raggedright\arraybackslash}X}
\begin{document}
\title{Right Numbers, Wrong Questions}
\subtitle{Evaluating LLMs on Underspecified Financial Queries}
\author{Radha Yadav}
\affiliation{\institution{TIFIN, India}\country{}}
\author{Prasanna Devadiga}
\affiliation{\institution{TIFIN, India}\country{}}
\author{Peeyush Mishra}
\affiliation{\institution{TIFIN, India}\country{}}
\author{Apoorv Singh}
\affiliation{\institution{TIFIN, India}\country{}}
\begin{abstract}
Financial assistants answer questions that sound factual but depend on what the user leaves unsaid: what a data figure measures, which product variant is meant, and the investor's situation and the rules that apply. An assistant that ignores these can still answer with a real number from its data, one that answers a different question. We call this \emph{frame collapse}. We study it in Indian mutual funds, where product variants and tax rules are formally defined, so the correct answer can be computed for every reading. One study tests each challenge. Given an ambiguously labelled figure, models used it, often under the user's label. Given every alternative labelled, they listed them all, but the layout decided which came first. Missing a fact about the investor, weaker agents assumed a default or an outdated tax rule; requiring the fact did not guarantee its use. Because such answers match their data, checks against the data pass them. An AI judge caught mislabelled figures only when told what each field means, and still passed answers that silently chose another reading. We show how these failures arise, not how often, and propose controls for financial assistants.
\end{abstract}
\maketitle

\section{Introduction}
Financial assistants answer questions that sound factual but rarely are. The answer to ``What is this fund's return?'' or ``How much will I get if I sell?'' depends on things the user does not say. The assistant must work these out before answering, and three things make that hard:

\begin{enumerate}
\item \textbf{Its data can be ambiguous.} A field named \texttt{growth} may hold a fund category's net inflow or its change in assets; a benchmark figure may come from the price index or the total-return index.
\item \textbf{Products come in variants.} Each mutual fund is sold in two \emph{plans}, Direct and Regular, with the same portfolio but different fees (Appendix~\ref{app:glossary}). Investors also put money in either as a lump sum or monthly. Each variant has its own return.
\item \textbf{The answer depends on the investor and the rules.} What a sale pays out depends on who sells, for example whether they live in India, and on the tax rule in force.
\end{enumerate}

Take ``What are the returns of HDFC Flexi Cap Fund?'' Over the same three years, its Direct plan returned 14.72\% a year and its Regular plan 13.97\%. Both figures are correct. Which one answers the user depends on the plan they hold, and the question does not say. On Rs.~10 lakh (Rs.~1 million), the gap is about Rs.~29,300.

\begin{figure*}[t]
\centering
\includegraphics[width=\textwidth]{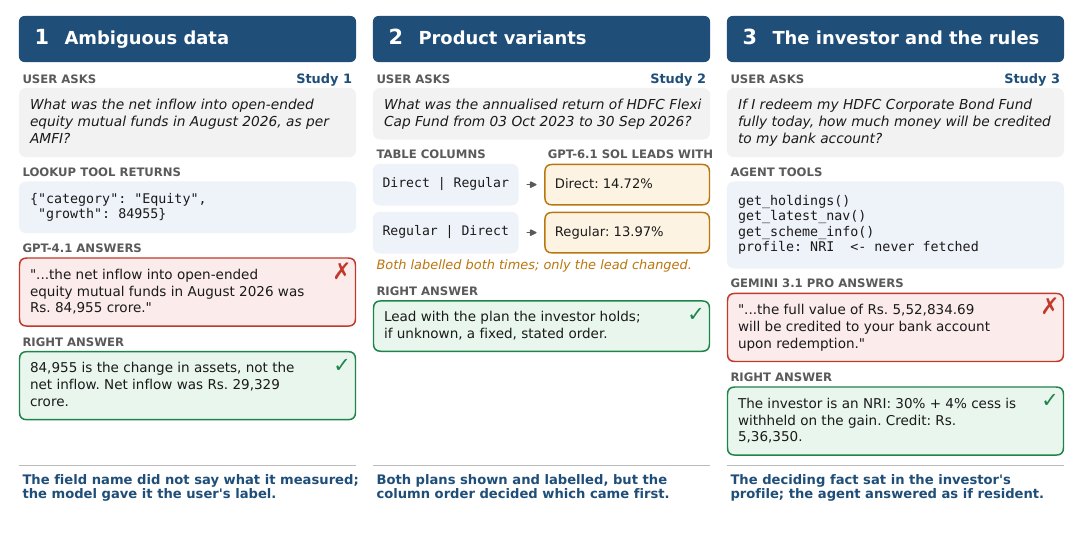}
\caption{One real case for each challenge. Model answers are quoted from the study logs, shortened with ellipses (currency shown as Rs.). (1) Study 1, a lookup tool returning one ambiguously named record. (2) One Study 2 question asked twice, with the table's plan columns in each order; each figure is that plan's three-year annualised return. (3) A Study 3 run of a model chosen because it failed in exploratory runs: a profile tool holding the investor's NRI status was available but never called. The panels illustrate each failure; Section~\ref{sec:results} reports how often it occurred, and Appendix Figure~\ref{fig:summary} summarises every measurement.}
\Description{Three panels, one per challenge. Panel 1, ambiguous data: asked for August 2026's net inflow into equity funds, a lookup tool returns a record with a field named growth of 84,955; GPT-4.1 reports a net inflow of Rs. 84,955 crore, but that figure is the change in assets and the net inflow was Rs. 29,329 crore. Panel 2, product variants: asked for HDFC Flexi Cap Fund's annualised return, GPT-6.1 Sol leads with the Direct plan's 14.72 percent when the Direct column comes first and with the Regular plan's 13.97 percent when the Regular column comes first, labelling both plans each time; the right behaviour is to lead with the investor's plan or a fixed, stated order. Panel 3, the investor and the rules: asked how much a full redemption of HDFC Corporate Bond Fund will credit to the bank, Gemini 3.1 Pro, which never fetched the investor's profile, answers Rs. 5,52,834.69; the investor is an NRI, so 30 percent plus 4 percent cess is withheld on the gain and the credit is Rs. 5,36,350.}
\label{fig:examples}
\end{figure*}

An assistant that leaves these questions unsettled still answers, with a real number that answers a different question. We call this \emph{frame collapse}: the model commits to one interpretation of an underspecified question without establishing or stating it. Unlike a numerical error or a hallucination, the figure is real and traceable to the evidence, so a check that verifies an answer against its evidence passes it.

We study frame collapse in Indian mutual funds. Plans, options and investing patterns are formally defined, NAV records are public, and tax rules depend on the investor and changed recently, so we can compute the correct answer for every reading. The setting matters in practice too: the regulator holds firms responsible for what their AI tools say~\cite{sebiai2025}. Each study tests one challenge, and Figure~\ref{fig:examples} shows a real case from each:

\begin{itemize}
\item \emph{Ambiguous data} (Study 1). Given one ambiguously named figure that was not the quantity asked for, seven of eight models used it, often under the user's label. Given every candidate, labelled, they rarely erred.
\item \emph{Product variants} (Study 2). Given both plans, labelled, most current models listed both, but the table's column order decided which came first.
\item \emph{The investor and the rules} (Study 3). Weaker agents assumed a default investor or an outdated tax rule. Requiring the fact got it fetched, but not always used. The newest models gathered it unprompted.
\item \emph{Checking.} An AI judge caught mislabelled figures only when told what each data field means, and still passed answers that silently chose another reading. A prompt that checks the data against the question cut wrong answers, mostly by asking the user back.
\end{itemize}

One description fits these results: what reaches the model's context sets the interpretation, and defaults fill the gaps. We formed it after the studies and did not test it, and the most careful current models were the clearest exceptions. The studies show how these failures arise, not how often, and AI models coded all answers, with a human check on a sample (Section~\ref{sec:limits}).

\textbf{Contributions.} (1) A taxonomy of how a financial question can be underspecified, with a materiality rule (Section~\ref{sec:problem}). (2) Three studies, one per challenge, with a computed reference for each reading (Sections~\ref{sec:design}--\ref{sec:results}). (3) Evidence on what a judge needs to catch these answers, and what a prevention check costs. (4) Controls for financial assistants, separating tested from proposed (Section~\ref{sec:discussion}). Appendix~\ref{app:glossary} is a glossary of Indian terms; Appendix~\ref{app:reporting} gives full counts, denominators and statistics.

\begin{table}[t]
\caption{Five defensible readings of ``What are the returns of HDFC Flexi Cap Fund?'', computed from AMFI NAV records through 30 September 2026, net of fund expenses, before exit load and tax. XIRR is the annualised return over dated cash flows (glossary, Appendix~\ref{app:glossary}). The IDCW (payout) figure is the investor's own cash-flow return, not a fund return; it exceeds Growth even though that investor ends with less (Appendix~\ref{app:terms}).}
\label{tab:hdfc}
\small
\begin{tabularx}{\columnwidth}{L{1.25}L{1.10}L{0.65}}
\toprule
\textbf{Plan, option, mode} & \textbf{Measure, period} & \textbf{Return} \\
\midrule
Regular, Growth & 3-year annualised & 13.97\% \\
Direct, Growth & 3-year annualised & 14.72\% \\
Regular, IDCW payout & 3-year XIRR, before tax & 15.12\% \\
Direct, Growth; Rs.~5,000 a month (SIP) & 5-year XIRR & 13.66\% \\
Direct, Growth; one lump sum & 5-year annualised & 15.67\% \\
\bottomrule
\end{tabularx}
\end{table}

\begin{table*}[t]
\caption{Fourteen dimensions along which a financial question can be underspecified.}
\label{tab:taxonomy}
\small
\begin{tabularx}{\textwidth}{L{0.39}L{1.50}L{1.11}}
\toprule
\textbf{Group} & \textbf{Dimensions} & \textbf{Example of an unsupported interpretation} \\
\midrule
A. Referent & A1 entity (fund, fund house, company, person, index; renamed schemes); A2 plan and option (Direct/Regular x Growth/IDCW); A3 unit and scale & ``HDFC'' read as HDFC Bank when HDFC's fund house was meant; a Regular-plan NAV given for a Direct holding \\
B. Measurement & B1 metric (absolute, annualised, XIRR; asset change vs net flow); B2 horizon and as-of date; B3 mode (lump sum vs SIP); B4 gross or net basis (before or after exit load, transaction taxes and tax) & A 3-year total return reported as a 3-year annualised return; a redemption credit quoted before tax \\
C. Evaluative & C1 comparator (total-return vs price index, category); C2 objective; C3 risk measure & ``Beat the Nifty'' judged against the price index \\
D. Contextual & D1 jurisdiction; D2 tax status and regime (including treaty relief); D3 market regime; D4 data and rule vintage & A capital-gains rule superseded in July 2024 applied to a 2026 sale \\
\bottomrule
\end{tabularx}
\end{table*}

\section{The Problem: What Must Be Settled Before Answering}
\label{sec:problem}
Table~\ref{tab:hdfc} extends the opening example: one question, five defensible readings, five correct figures. Each answers a different question.

\textbf{Dimensions.} A question has one appropriate answer only once its material dimensions are settled. Table~\ref{tab:taxonomy} groups fourteen dimensions: what is measured, how, against what, and in which context. We compiled the taxonomy from regulatory definitions and practice; it is not measured from real queries.

\textbf{Materiality and appropriate behaviour.} A dimension is \emph{material} when changing it moves the answer beyond the reporting tolerance; we score only material dimensions. The appropriate behaviour, in order: answer under the investor's known values; otherwise state the basis used and name the material alternatives (for an unknown plan, both plans, labelled); ask only when no basis exists; and decline when the data cannot answer. Investor-specific dimensions take three steps: gather the investor's facts, apply the rule to them, and use the rule version in force.

\textbf{Forms of failure.} Table~\ref{tab:forms} separates the failures we observe. Three are frame collapse: the answer commits to one value of a material open dimension, or attaches a measure to a figure, although nothing in the model's context settled it, and does not say so. The fourth, headline prioritisation, is weaker: both figures are shown and labelled, and only their order comes from the layout. The studies score these forms differently, so they do not yield one failure rate.

\begin{table*}[t]
\caption{The forms of failure, where each is tested and how it is scored. Forms (i), (ii) and (iv) are frame collapse; (iii) is a related, weaker presentation effect.}
\label{tab:forms}
\small
\begin{tabularx}{\textwidth}{L{0.62}L{1.10}L{1.18}L{0.55}L{0.55}}
\toprule
\textbf{Form} & \textbf{What the answer does} & \textbf{Example from our studies} & \textbf{Tested in} & \textbf{Scored by} \\
\midrule
(i) Mislabel & Attaches the user's label to a figure that measures something else & A month's change in assets reported as the net inflow & Study 1; judges & Codebook \\
(ii) Silent reading & Answers under another admissible reading without saying so & A lump-sum return given to an investor who invests monthly & Studies 1 and 2 & Codebook; extraction \\
(iii) Headline prioritisation & Shows and labels both readings, but the layout decides which leads & The first column's plan headlined & Study 2 & Code (order) \\
(iv) Omitted condition & Leaves out a decisive investor fact, or uses a superseded rule & An NRI's bank credit quoted before tax is withheld & Study 3 & Code, tool logs \\
\bottomrule
\end{tabularx}
\end{table*}

\section{Related Work}
\textbf{Ambiguity and clarification.} Ambiguous questions admit several correct answers~\cite{ambigqa}, and models rarely resolve them unprompted~\cite{clamber,laban2025lost}. Benchmarks test whether models ask for missing information~\cite{questbench,qian2024in3}, and asking can be learned~\cite{kuhn2022clam,andukuri2024stargate}. Conditional QA asks which conditions apply~\cite{sun2022conditionalqa}, and time-sensitive QA whether knowledge is current~\cite{vu2024freshqa}. Most financial benchmarks fully specify the question~\cite{finqa,financebench}; RealFin removes premises from exam questions~\cite{realfin}. Closest to us, FinInteract~\cite{fininteract} pairs default and intended readings of ambiguous financial questions and finds interaction, not financial knowledge, to be the bottleneck. Ambiguous database questions have reference answers for each interpretation~\cite{bhaskar2023ambiqt,saparina2024ambrosia}. We use underspecification for questions, not for model pipelines~\cite{damour2022underspecification}. We add computed references for each reading in a regulated domain, score the basis an answer states, and manipulate where the interpretation comes from, rather than only asking whether a model asks.

\textbf{Context, order and tools.} Irrelevant or counterfactual context misleads models~\cite{shi2023distracted,ming2025faitheval}. They favour options by position~\cite{zheng2024mcq,pezeshkpour2024order} and are sensitive to table and prompt format~\cite{sui2024table,sclar2024}. Agents over-rely on tools~\cite{yang2026overreliance,wu2024clasheval} and choose them by their descriptions~\cite{hasan2026mcpsmells}. Models accept questionable presuppositions~\cite{yu2023crepe,kim2023qa2} and defer to a user's stated view~\cite{sharma2024sycophancy}; a mislabel accepts the label the user's wording presupposes. Models also hold knowledge they do not use~\cite{knowingdoing,hager2024clinical}. Our context differs: it is not misleading, but true under another interpretation.

\textbf{Grounding and judges.} Answer checks usually test whether an answer is supported by its evidence, through attribution metrics~\cite{ais,factscore} or LLM judges~\cite{zheng2023judge}. Evaluator verdicts on underspecified queries change once user context is supplied~\cite{malaviya2025contexteval}. A frame-collapsed answer is supported by its evidence, so we test what else a judge needs, borrowing field descriptions from text-to-SQL~\cite{li2023bird}.

\section{Study Design}
\label{sec:design}
Each study tests one challenge by changing one thing the model sees (Table~\ref{tab:design}). Studies 1 and 2 leave out the investor; Study 3 adds the investor and the tax rule. Two further tests reuse Study 1's answers: can a judge catch the failure, and can a prompt prevent it?

\begin{table*}[t]
\caption{Design of the three studies. Each changes one thing that reaches the model's context. Cohorts and denominators: Appendix~\ref{app:cohorts}.}
\label{tab:design}
\small
\begin{tabularx}{\textwidth}{L{0.55}L{1.15}L{1.15}L{1.15}}
\toprule
 & \textbf{Study 1: ambiguous data} & \textbf{Study 2: product variants} & \textbf{Study 3: the investor and the rules} \\
\midrule
What changes & No data, a thin record or labelled records, returned by a lookup tool; precise or vague wording & Plan or mode open or stated; Direct or Regular column first (counterbalanced); table in the prompt or tools & Whether the agent's interface mentions the decisive fact, demands it, or supplies the current rule \\
Models & GPT-4.1, Gemini 2.5 Flash, Qwen3.8 27B; replicated on five more & Eleven current models & Gemini 3.1 Pro and GPT-4.1, selected for failing; five further models as checks \\
Scale & 22 questions; 1,188 answers & 148 items; 3,256 answers & Two redemption cases; 350 planned runs, 80 check and final runs \\
Outcome & Wrong answer; mislabel & Silent on the open dimension; which plan leads & Fetched the fact; figure within tolerance \\
Scored by & An AI coder with a fixed codebook & Code, from two AI extractors' readings & Code and tool logs \\
Read with care & Codebook stricter than the definition; illustrative figures & Rates over answers both extractors agree on; extractors are tested models & Two cases; models chosen for failing; mechanisms, not rates \\
\bottomrule
\end{tabularx}
\end{table*}

\textbf{Study 1: ambiguous data.} We wrote 22 questions in three wordings: precise, vague, and a vague paraphrase. Each came with no data, a \emph{thin} record (one record whose field name does not say exactly what its figure measures) or \emph{labelled} records (every candidate, each labelled). The data appeared as a lookup tool's output under the instruction `If data is provided, use it'. Scheme figures are illustrative. We also ran each item with the question rewritten to name the field it needs. Because the three models are older, we replicated the thin-data test on GPT-6.1 Sol, Gemini 3.8 Flash and three small open-weight models (Llama 3.1 8B, Mistral Nemo 12B, Qwen3 8B).

\textbf{Study 2: product variants.} Eleven current models (Claude Opus 5.5, GPT-6.1 Sol, GPT-6 Astra, Gemini 3.8 Flash, Grok 4.7, Kimi K3, GLM-5.3, DeepSeek V4.1 Flash, Qwen3.5-122B, Llama 4 Maverick, gpt-oss-120b) answered questions on real AMFI NAV records for ten funds. They gave either a fund's three-year return, with the plan open or stated, or the return of an investor ``investing since 2021'', with the mode open or stated. We counterbalanced the order of the Direct and Regular columns, in a table or in a tool's search ranking.

\textbf{Study 3: the investor and the rules.} An agent with tools for the investor's profile, holdings, NAVs and scheme facts answered two redemption questions, each turning on one profile fact (Table~\ref{tab:cases}). Each arm changes one element from its comparator: a \emph{mention} (a tool description only), a \emph{demand} (a tool argument or one-line checklist requiring the fact), or \emph{rule supply} (the current rule as a tool or prompt text). Each arm ran at least five times per case.

\begin{table*}[t]
\caption{The two Study 3 cases. Neither question mentions tax status or other gains. Both redeem at the 30 September 2026 NAV and assume no surcharge, a PAN on file, no treaty relief and no exit load. The NRI reference is the bank credit after TDS, not final tax; the resident reference is final tax, assuming other income above the basic exemption and no carried-forward losses. Rounding conventions: Appendix~\ref{app:terms}.}
\label{tab:cases}
\small
\begin{tabularx}{\textwidth}{L{0.70}L{1.35}L{1.45}L{0.65}L{0.85}}
\toprule
\textbf{Study 3 case} & \textbf{Investor and holding} & \textbf{Rule that decides it} & \textbf{Reference answer} & \textbf{If the fact is missed} \\
\midrule
Non-resident withholding & NRI; Rs. 5 lakh in a debt fund (HDFC Corporate Bond) bought January 2025; gain Rs. 52,835 & Debt funds (since April 2025, funds with more than 65\% in debt and money-market instruments) bought on or after 1 April 2023 are taxed at slab rate, whatever the holding period; for an NRI, 30\% plus 4\% cess is withheld at source & Bank credit Rs. 5,36,350 & Rs. 5,52,835 (overstated by Rs. 16,484) \\
Booked gains & Resident; Rs. 5 lakh in an equity fund (Parag Parikh Flexi Cap) bought June 2023; Rs. 60,000 of long-term gains on listed shares already booked this tax year & Long-term gains on listed equity and equity-oriented fund units (held over 12 months) taxed at 12.5\% plus cess above a Rs. 1.25 lakh annual exemption shared with other such gains (before 23 July 2024: 10\% above Rs. 1 lakh) & Gain Rs. 2,66,118; tax Rs. 26,145 (tax year 2026-27) & Rs. 18,345 if the exemption is not shared; Rs. 23,516 under the superseded rule \\
\bottomrule
\end{tabularx}
\end{table*}

\textbf{Checks and prevention.} AI judges scored Study 1's 162 thin-data answers, with and without the data, a neutral data dictionary and the user's intended reading. A \emph{data-match check}, a prompt that states what the question asks for and what each field measures before answering, ran on Study 1's thin items with one open-weight model.

\textbf{Measurement.} Code or a fixed codebook decides correctness; models only extract or code what an answer did. References for Studies 2 and 3 are computed in code from AMFI NAV records and the Income-tax Act, 2025. Three limits apply throughout. One AI coder applied Study 1's codebook, checked by blind recodes, a coder from another model family, a model-free text rule and, on 50 answers, a finance-trained human coder. Study 2's two extractors (Claude Opus 5.5 and GPT-6.1 Sol) are also tested models, and its rates count only answers both agree on. Study 3 scores the figure, not its disclosure, and uses models chosen because they failed in exploratory runs. Appendices~\ref{app:terms}--\ref{app:stats} give tolerances, coding checks and statistics.

\section{Results}
\label{sec:results}
The results follow the three challenges of Figure~\ref{fig:examples}, with Study 3 split into a missing fact and a demanded fact. Sections~\ref{sec:checks} and~\ref{sec:prevent} ask whether such answers can be caught or prevented. Appendix Figure~\ref{fig:summary} summarises every measurement.

\subsection{Ambiguous data: one figure in context is adopted (Study 1)}
\label{sec:one}
Given one figure, models committed to it. On the same 14 questions and three wordings, the three Study 1 models gave 7 wrong answers of 126 with labelled data. When a lookup tool returned a thin record whose figure was not the quantity asked for, they gave 96 of 126. Only 1 of the 162 thin-data answers asked the user anything.

The commitment took two forms. A mislabel attaches the user's label to the figure. Asked for the net inflow and given \texttt{\{"category": "Equity", "growth": 84955\}}, a month's change in assets, GPT-4.1 replied that ``the net inflow ... was Rs. 84,955 crore'': 2.9 times the actual net inflow of Rs. 29,329 crore. On precise wording, 26 of 42 thin answers carried the user's label; none of the 42 labelled answers did. A silent reading is true under another interpretation but does not say so. AI coders disagree on how the 96 split between the two forms, but agree on whether an answer is wrong (kappa 0.70--0.95 across three coders). A finance-trained human coder who blind-recoded a sample of 50 answers agreed with the codebook on both (kappa 0.90 on wrong versus not; 33 of 34 on the form).

A mislabel can reverse a decision. All nine thin answers to one question said the large-cap category beat the Nifty 50, citing 7.6\%; three called this the total-return index (TRI), SEBI's required benchmark since 2018. But 7.6\% was the price index's return. The category's 8.2\% beat the price index, not the TRI's 8.9\%.

The pattern held across models, with one exception. Of five further models, blind-coded on the same questions, four were wrong on 99 to 113 of 126 answers. GPT-6.1 Sol was wrong on 38 and mostly flagged the gap or asked. Overall, seven of eight models were wrong in 69--90\% of these answers (Table~\ref{tab:repl}). Attaching the label partly reflects a reasonable presupposition that the tool returned what was asked~\cite{yu2023crepe}.

Accurate field names helped only where the record named a different thing. Thin and labelled records differ in naming and in whether the asked figure is present. A planned arm on the three small models separated the two: it kept the single record but renamed its fields to say what each measures (\texttt{aum\_change\_cr} instead of \texttt{growth}). Mislabels fell from 42 to 10 of 54 where the record now named a different company, scheme, plan or method. Where it named a close relative of the asked measure, such as asset change for net inflow, they barely moved (59 to 57 of 72): the user's wording set the label. The overall effect is suggestive, not confirmatory (Appendix~\ref{app:study1}).

\subsection{Product variants: set out, but in layout order (Study 2)}
\label{sec:labelled}
Capability was not the problem: with plan and mode specified, Claude Opus 5.5 and GPT-6.1 Sol were right on every Study 2 item, and most models on 75--99\%.

Given every candidate, labelled, most models set out the alternatives. Study 1's vague questions with labelled data were flagged or asked in 82 of 108 answers. In Study 2, Claude Opus 5.5 and GPT-6.1 Sol named the open plan or mode in every agreed answer. Five models (Gemini 3.8 Flash, Grok 4.7, Qwen3.5-122B, gpt-oss-120b, Llama 4 Maverick) gave 140 of the 149 silent agreed answers, out of 713 agreed open-item answers that stated a figure (Figure~\ref{fig:study2}a--b; coding exceptions in Appendix~\ref{app:study2}).

The layout, however, decided which came first. Nothing in the question or records says which plan to list first. In 130 of 145 answers that showed both plans, the first-listed plan was the first column's. Of the ten models with such answers, nine followed the first column more often and one tied (model-level sign test p = 0.004). Swapping the columns flipped the leading plan in 61 of 84 matched pairs (Figure~\ref{fig:study2}c). Nothing was hidden: both figures were present and labelled. The layout set the priority; we did not test whether readers act on the first figure.

\begin{figure*}[t]
\centering
\includegraphics[width=0.95\textwidth]{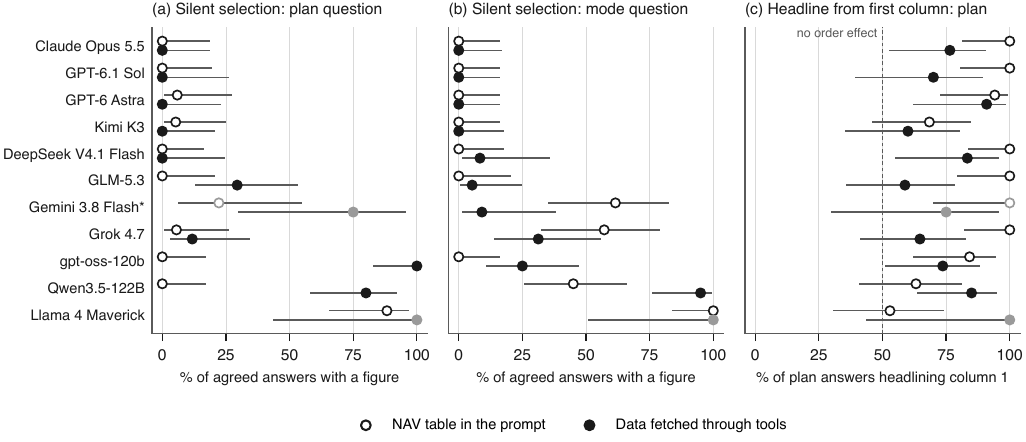}
\caption{Study 2, per model, over agreed answers that stated a figure. (a, b) Silent selection (form ii, silent reading): answered without naming the open plan or mode, or asking. (c) Share of plan answers headlining the first column's plan (order counterbalanced; 50\% = no order effect); most of these answers show both plans. Open: NAV table in the prompt; filled: data fetched through tools; bars: Wilson 95\% intervals; grey: fewer than 10 answers. *Gemini 3.8 Flash's plan cases are side-by-side layouts that an audit coded as flagged.}
\Description{Three dot-and-interval panels, one row per model. Panels a and b show silent selection rates for the plan and mode questions: near zero for Claude Opus 5.5, GPT-6.1 Sol, GPT-6 Astra, Kimi K3 and DeepSeek V4.1 Flash, and high for some of Gemini 3.8 Flash, Grok 4.7, gpt-oss-120b, Qwen3.5-122B and Llama 4 Maverick. Panel c shows the share of plan answers headlining the first column, mostly between 60 and 100 percent, to the right of the 50 percent no-order-effect line.}
\label{fig:study2}
\end{figure*}

\subsection{The investor and the rules: a missing fact is filled by a default (Study 3)}
\label{sec:missing}
When a needed fact was missing from context, a default filled the gap. With residency removed from its profile, Gemini 3.1 Pro gave the resident (no-withholding) figure in 20 of 20 runs and never flagged the gap. In Study 2, all 81 agreed silent mode answers assumed a lump sum, although the investor wrote ``I have been investing... since 2021'' and the table showed no cash flows. GPT-4.1 and GPT-4.1-mini, with a knowledge cutoff before India's July 2024 Budget, applied the capital-gains rule it superseded. These defaults are consistent with model priors.

The omission did not transfer to frontier agents. In final planned runs, GPT-6.1 Sol and Gemini 3.8 Flash fetched the fact unprompted and answered both cases correctly in all 20 runs. Gemini 3.5 Flash also gathered it unaided (10 of 10), and GPT-5.5 showed no gap. Study 3 thus shows a mechanism in some models, not a general rate.

\subsection{A demanded fact is fetched, but not always used (Study 3)}
\label{sec:demand}
\begin{figure*}[t]
\centering
\includegraphics[width=0.78\textwidth]{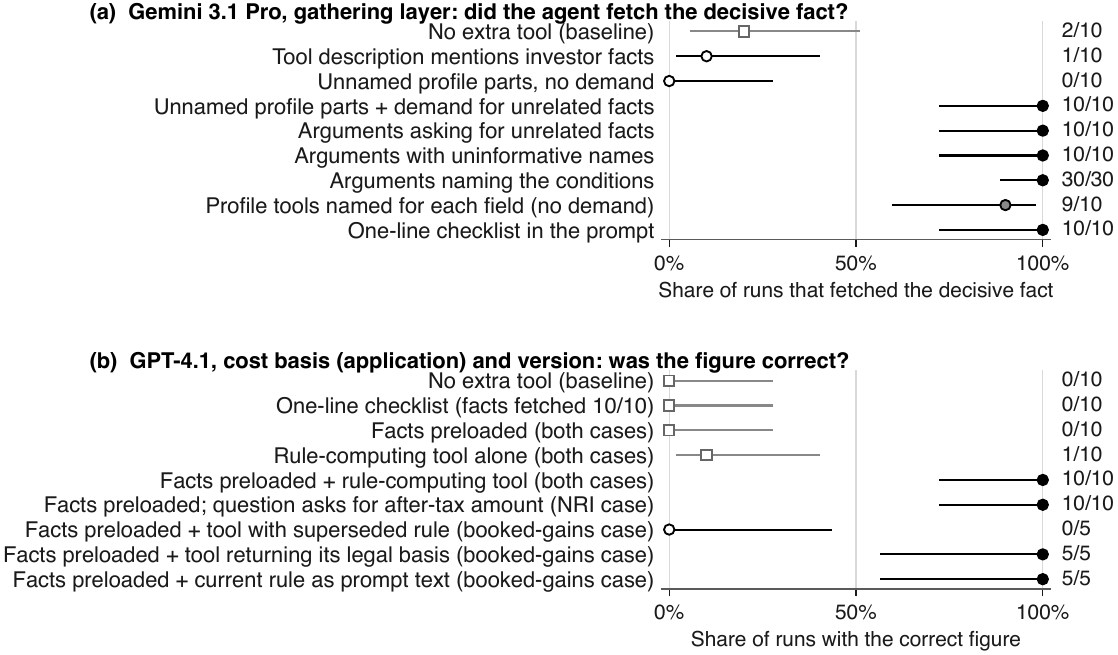}
\caption{Study 3 arms. (a) Gemini 3.1 Pro: share of runs that fetched the decisive fact. (b) GPT-4.1: share of runs with the correct figure. Filled circles: a demand or fix; open circles: a mention, no demand or a stale rule; grey squares: no fix; grey circle: field-named tools without a demand. Lines: Wilson 95\% intervals; five runs per case, two cases per arm unless a case is named.}
\Description{Two dot-and-interval panels. Panel a: Gemini 3.1 Pro fetched the decisive fact in 2 of 10 runs at baseline, 1 of 10 with a mention and 0 of 10 with unnamed profile parts, but in 10 of 10 to 30 of 30 runs with any demand, and 9 of 10 with profile tools named for each field. Panel b: GPT-4.1 gave the correct figure in 0 of 10 runs at baseline, with a checklist or with facts preloaded, 1 of 10 with a rule tool alone, 10 of 10 with facts plus a rule tool or when the question asked for the after-tax amount, 0 of 5 with a superseded-rule tool, and 5 of 5 with the legal basis or the current rule as prompt text.}
\label{fig:evidence}
\end{figure*}

A demand, unlike a mention, made a failing agent fetch the fact. With a demand, as a tool argument or a one-line checklist, Gemini 3.1 Pro fetched the decisive fact in 10 of 10 runs. With only a mention in a tool description, it did so in 1 of 10 (one-sided Fisher p = 6.0 $\times$ 10$^{-5}$; Figure~\ref{fig:evidence}a). When the demand named the decisive fact, it answered both cases correctly in every run. A demand for unrelated facts got the profile fetched, but the non-resident case right in at most 1 of 5.

Fetching did not ensure use. Gemini 3.1 Pro fetched the investor's NRI status in 15 of 15 runs, yet repeated a compute tool's resident-based figure in 14. GPT-4.1 fetched the fact under a checklist in 10 of 10 runs and was correct in none (Figure~\ref{fig:evidence}b), for a different reason in each case. In the non-resident case, knowledge was not the gap: GPT-4.1 stated the 30\% withholding rate when asked, yet quoted the bank credit before tax in 10 of 10 agent runs. It applied the withholding only when the question explicitly asked for the amount after tax (10 of 10). In the booked-gains case, it consistently produced the superseded rule. Supplying the current rule as prompt text fixed this (5 of 5 against 0 of 5); a tool computing tax under the superseded rule was relayed in 5 of 5.

\subsection{Can a check see it?}
\label{sec:checks}
A frame-collapsed answer agrees with its own data, so a judge that checks only support can accept it, as the answering model did. We gave two judges Study 1's 162 thin answers and varied what they saw (Table~\ref{tab:judges}).

The data alone did not help; the data's meaning did. Told the intended reading and shown the data, GPT-6.1 Sol named 17 of 44 mislabels. With a neutral data-dictionary entry for each field, written from the source record without reference to the question, it named 40 of 44 (9 of 13 questions improved, none worsened; sign p = 0.004). False alarms on acceptable answers rose from 15 to 23 of 61.

The gain held without the intended reading. A deployed judge does not know what the user meant, so planned arms removed the reading and reworded the rubric to `answers the question as asked'. With the dictionary, Sol still named 40 of 44 mislabels, against 23 with the data alone, with 17 false alarms of 61.

Silent readings, however, passed. In that condition Sol passed 42 of the 52 silent readings of vague questions, while rating every mislabel wrong. A silent reading does answer a vague question as asked; it lacks only a stated interpretation. The dictionary catches mislabels (i). Catching silent readings (ii) would need a rubric that requires a stated interpretation, or a reference for each reading; we tested neither. As a release gate, the dictionary judge would also reject about a third of acceptable answers. The small open-weight judge behaved alike (Appendix~\ref{app:judges}).

\begin{table*}[t]
\caption{GPT-6.1 Sol as a judge of Study 1's 162 thin answers: 44 mislabels, 96 wrong answers (26 precise, 70 vague) and 61 acceptable (right, flagged or asked) answers. Both judges, all arms and tests: Appendix~\ref{app:judges}.}
\label{tab:judges}
\small
\begin{tabularx}{0.8\textwidth}{L{0.62}L{1.28}L{1.02}L{1.02}L{1.06}}
\toprule
\textbf{Told reading} & \textbf{Sees} & \textbf{Mislabels named (of 44)} & \textbf{Wrong rated correct (of 96)} & \textbf{Acceptable rejected (of 61)} \\
\midrule
Yes & Question & 14 & 18 & 15 \\
Yes & + data & 17 & 43 & 15 \\
Yes & + data + dictionary & 40 & 0 & 23 \\
No & + data & 23 & 53 & 15 \\
No & + data + dictionary & 40 & 42 & 17 \\
\bottomrule
\end{tabularx}
\end{table*}

\begin{table*}[!t]
\caption{Controls, the failure each addresses, and the evidence here. Rows 1--4 have evidence on few models and cases; rows 5--8 are proposed and untested. Full table with Indian implementations: Appendix~\ref{app:controls}.}
\label{tab:controls}
\small
\begin{tabularx}{\textwidth}{L{0.15}L{1.75}L{0.35}L{1.75}}
\toprule
\textbf{\#} & \textbf{Control} & \textbf{Form} & \textbf{Evidence here} \\
\midrule
1 & Take each decisive investor fact as a required, typed input from the records & (iv) & Made both failing agents fetch it (GPT-4.1 via a checklist); fixed one; frontier agents did not need it \\
2 & Supply the rule in force from a dated source; compute tax in code & (iv) & Current rule fixed GPT-4.1 (5 of 5 vs 0 of 5); a stale rule tool was relayed \\
3 & State what each data field measures; check it against the question & (i) & Helped where the record named a different entity (exploratory); prompt check over-asked \\
4 & Evaluate with the data's meaning; add a reference for each reading & Judges & Dictionary tested; per-reading references untested \\
5 & Label every figure with the interpretation used & All & Proposed \\
6 & Order alternatives by the user's own holding & (iii), (ii) & Proposed; order effect observed \\
7 & Keep servicing separate from advice & Advice & Proposed \\
8 & Log interpretation, facts and rule version & Audit & Proposed \\
\bottomrule
\end{tabularx}
\end{table*}

\subsection{Can it be prevented, and at what cost?}
\label{sec:prevent}
Section~\ref{sec:demand} showed that, for the agents selected for failing, a demand repaired fetching and a supplied current rule repaired the rule version. On the answer side, we tested a data-match check on Study 1's thin items with one open-weight model (gpt-oss-20b).

The check cut wrong answers, mostly by asking. Wrong answers fell from 38 to 9 of 53 (14 of 18 questions improved, none worsened), and appropriate behaviour rose from 12 to 29. But questions back to the user rose from 1 to 33 of 53, and plainly correct answers fell from 11 to 0. On labelled data, where no question was needed, it still asked in 21 of 53. If an unnecessary question costs a fraction of a wrong answer, the check beats both doing nothing and always asking only when that fraction is between a half and about one (Appendix~\ref{app:datamatch}). It needs a rule that lets the model answer when the data match.

Cheaper interventions showed no effect. Mentioning the facts in a tool description did not make the failing agent fetch them (1 of 10 against 2 of 10 at baseline), though ten runs per arm cannot exclude a moderate effect. Rewriting Study 1's questions to name the needed field left wrong answers unchanged (131 to 130 of 558 paired answers). Removing the instruction `If data is provided, use it' did not reduce mislabels.

\section{Discussion}
\label{sec:discussion}
\subsection{What the studies show together}
One description fits the three studies: the interpretation a model adopts tracks what reaches its context, and defaults fill what the context leaves open. We formed it after the studies. A single candidate is committed to; labelled candidates are set out in layout order; a missing fact is filled by a default; a demanded fact is fetched, but not always used.

Three manipulations could have contradicted it and did not. Labelled data removed mislabels (64 of 66 precise answers right). Swapping the columns flipped the leading plan (61 of 84 pairs). A demand, unlike a mention, made a failing agent fetch the fact (10 of 10 against 1 of 10).

Three limits bound it. First, context is not sufficient: a retrieved fact was not always used, and on close measures the user's wording outweighed an accurate record. Second, it depends on the model. GPT-6.1 Sol still mislabelled 10 of 42 precise answers; Claude Opus 5.5 was tested only in Study 2; and the agent omissions did not transfer to GPT-6.1 Sol or Gemini 3.8 Flash. In the most careful current models, only layout order and, for judges, silent readings persisted. Third, when a model chooses its own context, as the frontier agents did, the description predicts nothing in advance.

Two models were tested in all three settings. GPT-6.1 Sol mostly flagged or asked in Studies 1 and 2 and answered every Study 3 run correctly. Gemini 3.8 Flash mislabelled (27 of 42 precise answers) and gave silent readings on most tabled mode questions (8 of 13), yet fetched the missing fact in every Study 3 run. This fits the description, but because the studies also differ in task, format and scoring, it does not show that context is the cause.

\subsection{Controls for a financial assistant}
The studies point to one principle: settle the material dimensions against the question, the data and the investor's records before answering, and label the answer with the interpretation used. Table~\ref{tab:controls} turns this into controls, separating tested from proposed. The tested parts were not tested together.

By money at stake, mislabels (a verdict reversed, a flow overstated 2.9-fold) and omitted investor conditions (a credit overstated by Rs. 16,484 on one redemption) matter most in our cases; we did not test headline prioritisation on readers. SEBI makes regulated firms responsible for their AI tools' outputs~\cite{sebiai2025}, and execution-only platforms may not give advice~\cite{sebieop2023}. The controls apply in any jurisdiction but were tested only in India; other markets have analogues, such as the UK's shared capital-gains exemption. Appendix~\ref{app:controls} gives Indian implementations and the full evidence.

\section{Limitations}
\label{sec:limits}
\textbf{Coding.} AI models did all coding (Study 1) and extraction (Study 2); a finance-trained colleague who is not an author checked a sample. On 50 answers enriched for wrong ones, the human, a model-free rule, blind coders and a coder from another model family agree on whether an answer is wrong (kappa 0.70--0.95; the human 0.90). The other-family coder recoded 10 of 15 silent readings as mislabels, but the human kept all 15, which supports the split between forms (i) and (ii), though on one coder and 50 answers. Study 2's extractors are also tested models.

\textbf{Samples.} Study 1's figures are illustrative and its main models older; the replications bound this. The accurate-record arm ran on three small models. Study 3's models were selected for failing, and its arms ran five times on two cases; its tolerance admits an omitted cess. The judge arms rest on 162 answers.

\textbf{Scope.} One jurisdiction and product family. We did not measure how often real users leave each dimension open, so nothing here estimates prevalence. Appendix~\ref{app:limits} discusses what extending the studies beyond India would show, and lists further limits.

\section{Conclusion}
A number can be present in an assistant's evidence, correct under some reading, and still answer a question the user did not ask. In our studies, models often neither established nor stated the interpretation they adopted; it came from what was nearest: a field name, a column order, a default about mode, residency or tax. Because such answers trace to their data, a check of support passes them, and a judge recognised most mislabels only when told what each field measures. Benchmarks should therefore ship a computed reference for every admissible reading and score whether a model established the reading, not just the number. Judges should be tested without the intended reading, which a deployed judge never has. The next test is the proposed combination of required investor facts, dated rules and labelled interpretations, evaluated together on real user queries with human coding; it would also estimate how often frame collapse occurs. Tax, insurance, banking and health questions share this structure of regulated variants and personal conditions, and are natural next domains.

\bibliographystyle{ACM-Reference-Format}
\bibliography{references}

\appendix
\section{Glossary of Indian Terms}
\label{app:glossary}
Table~\ref{tab:glossary} defines the Indian mutual-fund and tax terms used in the paper. Three words name three different things: a \emph{plan} is Direct or Regular, an \emph{option} is Growth or IDCW, and a \emph{mode} is a lump sum or an SIP.

\begin{table*}[t]
\caption{Glossary of Indian mutual-fund and tax terms.}
\label{tab:glossary}
\small
\begin{tabularx}{\textwidth}{L{0.45}L{1.55}}
\toprule
\textbf{Term} & \textbf{Meaning} \\
\midrule
Plan: Direct, Regular & Every scheme is offered in two plans that hold the same portfolio and differ only in where the investor bought it. A Direct plan is bought without a distributor (from the fund house, an execution-only platform or an adviser). A Regular plan is bought through a distributor, and its expense ratio includes the distributor's commission. The two plans therefore have different NAVs and returns. \\
Option: Growth, IDCW & Within each plan, the Growth option pays nothing out, so gains stay in the NAV. The income distribution cum capital withdrawal (IDCW) option pays out distributions, which can include the investor's own capital; its NAV falls with each payout. \\
Mode: lump sum, SIP & How the money went in: one single investment, or a systematic investment plan (SIP) that invests a fixed amount every month. \\
NAV & Net asset value per unit: the daily price at which units are bought and redeemed. \\
Expense ratio; exit load & The annual fee taken from the fund's assets (already reflected in the NAV); a fee charged on redemption within a set period. \\
Annualised return; XIRR & Annualised return: the return per year between two dates for a single investment. XIRR: the annualised return computed over dated cash flows, as in an SIP or IDCW payouts (an internal rate of return). \\
AUM; net inflow & Assets under management. Net inflow is money invested minus money redeemed; a change in AUM also includes market movements, so the two differ. \\
Price index; TRI & A price index tracks prices only; a total-return index (TRI) also reinvests dividends. SEBI has required funds to be benchmarked against a TRI since 2018. \\
Nifty 50 & An index of fifty large Indian companies. \\
Equity-oriented fund; debt fund & Tax classes. Long-term gains (held over 12 months) on listed equity and equity-oriented fund units are taxed at 12.5\% above an annual exemption. Gains on debt funds (since April 2025, funds with more than 65\% in debt and money-market instruments) bought on or after 1 April 2023 are taxed at the investor's slab rate, whatever the holding period. \\
Slab rate & The investor's income-tax bracket rate. \\
Long-term gains exemption & Rs.~1.25 lakh per tax year (Rs.~1 lakh before 23 July 2024), shared across all long-term gains on listed equity and equity-oriented fund units in that year. \\
NRI & Non-resident Indian: an investor not resident in India for tax purposes. \\
TDS; bank credit & Tax deducted at source: tax withheld before redemption proceeds reach the investor's bank. For an NRI's debt-fund gain, 30\% plus cess. The bank credit is the amount actually paid into the bank, after any TDS. \\
Cess & A 4\% levy charged on top of the tax. \\
PAN; folio & Permanent Account Number, India's tax identifier; an investor's account with a fund house. \\
Tax year & The April--March year; tax year 2026-27 runs from April 2026 to March 2027. \\
July 2024 Budget; Income-tax Act, 2025 & The 2024 Budget raised the long-term rate on equity gains from 10\% to 12.5\% and the exemption from Rs.~1 lakh to Rs.~1.25 lakh, from 23 July 2024. The Income-tax Act, 2025 is the law in force from 1 April 2026. \\
Lakh; crore & One lakh is 100,000 and one crore is 10 million. Indian digit grouping writes 536,350 as 5,36,350. \\
SEBI; AMFI & The Securities and Exchange Board of India, the market regulator; the Association of Mutual Funds in India, the industry body whose records give each fund's NAVs. \\
Execution-only platform & A platform that executes transactions in Direct plans and may not give advice. \\
\bottomrule
\end{tabularx}
\end{table*}

\section{Detailed Reporting}
\label{app:reporting}
This appendix reports the counts, denominators, coding checks and statistics behind the main text, which states only what the argument needs. Nothing here changes a main-text claim.

\subsection{Terms, tolerances and conventions}
\label{app:terms}
\textbf{Terms.} A \emph{reference answer} (gold) is computed in code. \emph{Precise} wording fully specifies the question; there the open element is what the data field measures. A \emph{mislabel} attaches the user's specific label to a figure that measures something else. A \emph{silent reading} answers under another interpretation without saying so and without a false label (form ii; the codebook's `wrong reading'); \emph{wrong} means a mislabel or a silent reading. A \emph{flag} names the unresolved dimension or the assumption used. \emph{Agreed} answers are those on which both Study 2 extractors agree; answers they disagree on are left out of agreed rates.

\textbf{Departures from the definition.} Study 1's codebook is stricter than the definition: it counts an answer that names the figure's true measure without naming the alternative as wrong, so relative to the definition its counts lean high (a blind recode confirmed every sampled wrong code but cannot show how many wrong answers were missed). On wrong versus not, a coder from another model family, gpt-oss-20b, agreed at kappa 0.82, two blind Claude coders at 0.70 and 0.95, and a human coder at 0.90. Study 2 counts labelled side-by-side answers as flagged; five unworded Gemini 3.8 Flash layouts, which an audit codes as flagged, are kept as silent and marked in Figure~\ref{fig:study2}. Study 3 scores the figure, not the disclosure.

\textbf{Tolerances.} The reporting tolerance is 5\% of the reference (at least 0.5 points) in Study 2 and 0.5\% (credits) or 8\% (tax) in Study 3, with floors of Rs. 100 and Rs. 200. The Study 3 tolerances admit an omitted 4\% cess but not a missed withholding (3.1\% of the credit), an unshared exemption (30\% below the tax reference) or the superseded rule (10\% below). In Study 2 the other plan's figure lies outside the tolerance for all ten funds, but only narrowly for HDFC Flexi Cap Direct (gap 0.75, tolerance 0.74 points), so plan is scored from the answer's stated label rather than its value; a 0.25-point tolerance would change 28 of 1,591 specified-item scores.

\textbf{Table~\ref{tab:hdfc}.} Each return period starts at the first NAV published on or after its start date; fund factsheets instead use the last NAV before a non-business start date. The IDCW figure is the payout investor's money-weighted outcome: it exceeds Growth, yet that investor receives less in payouts plus final holding value (Rs. 144.1 against Rs. 147.9 per Rs. 100 invested, before the slab-rate tax on payouts). Not a valid alternative: reading the Regular IDCW NAV as though it were the Growth NAV gives 3.62\%, a misreading. Direct plans are bought without a distributor (from the fund house, an execution-only platform or an adviser); the NAV records are published through the Association of Mutual Funds in India (AMFI), the industry body. Across ten funds, the Direct-Regular three-year difference on Rs. 10 lakh is Rs. 21,800--48,300.

\textbf{Currency and Table~\ref{tab:cases}.} One lakh is 100,000 and one crore 10 million; Indian digit grouping writes 536,350 as 5,36,350. The NRI overstatement of Rs. 16,484 is about USD 190 at Rs. 87 per dollar. Figures are rounded to the rupee separately (the overstatement is computed before rounding); stamp duty on purchase (about Rs. 25) is ignored.

\subsection{Evidence summary}
\begin{figure*}[t]
\centering
\includegraphics[width=0.95\textwidth]{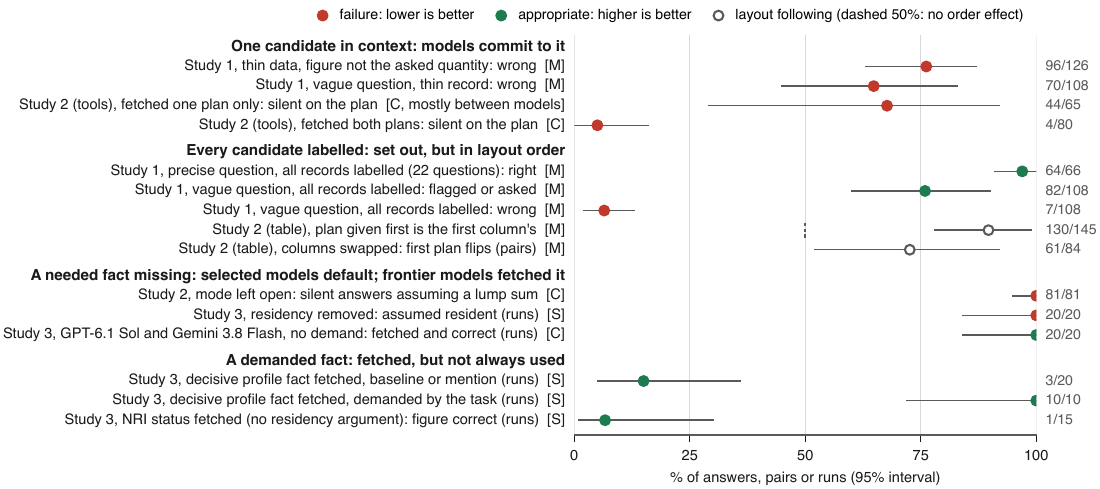}
\caption{Evidence summary: what reached the model's context, and what it did. Each block is one situation; each line is one measurement with its count and 95\% interval. Red: failures (lower is better); green: appropriate behaviour (higher is better); open circles: share following the layout (dashed line at 50\%: no order effect). [M] a manipulation; [C] a comparison across models or answers; [S] runs of a model chosen because it failed in exploratory runs. Lines are comparable within a block, not across blocks: the studies use different models and scoring rules (Section~\ref{sec:design}). Populations and interval methods: Appendix~\ref{app:cohorts}.}
\Description{A dot-and-interval chart in four blocks. Block one, one candidate in context: high failure rates with a thin record (96 of 126 wrong, 70 of 108 vague answers wrong) and 44 of 65 silent when only one plan was fetched, against 4 of 80 when both were. Block two, every candidate labelled: 64 of 66 precise answers right, 82 of 108 vague answers flagged or asked, 7 of 108 wrong; the first column's plan led in 130 of 145 answers and swapping columns flipped it in 61 of 84 pairs. Block three, a needed fact missing: 81 of 81 silent mode answers assumed a lump sum; a selected agent assumed residency in 20 of 20 runs; two frontier agents fetched the fact and were correct in 20 of 20. Block four, a demanded fact: fetched in 3 of 20 runs at baseline or with a mention, 10 of 10 when demanded; with the residency fact fetched but not passed to the compute tool, the figure was correct in 1 of 15.}
\label{fig:summary}
\end{figure*}

Figure~\ref{fig:summary} collects the main measurements behind Section~\ref{sec:results}, grouped by what reached the model's context.

\subsection{Cohorts and denominators}
\label{app:cohorts}
\textbf{Study 1.} Each run is 22 questions x 3 wordings x 3 data conditions x 3 models = 594 answers; with the explicit-field restatement run, 1,188 in all. The denominator ladder for the original arm: 594 generated $\rightarrow$ 558 included, after excluding the 36 thin-data answers to four questions whose thin label was itself wrong $\rightarrow$ 162 with thin data (18 questions) $\rightarrow$ 126 whose figure was not the asked quantity (14 questions) $\rightarrow$ 42 of those on precise wording. All 96 wrong thin answers fall among the 126. Figure~\ref{fig:summary}'s Study 1 vague lines use the 18 questions with a usable thin record. Original and restatement arms are reported separately.

\textbf{Study 2.} 148 items x 11 models x table and tools = 3,256 answers. Silence rates condition on extractor agreement and on the answer stating a figure (713 agreed open-item answers stated one). Specified-item accuracy is over specified items, not over the whole 148-item inventory.

\textbf{Judges.} The 162 thin answers partition into 96 codebook-wrong (26 precise, 70 vague; 44 of them mislabels, 26 precise and 18 vague), 61 right, flagged or asked, and 5 coded as date confusion.

\textbf{Data-match.} Headline counts (38 to 9 of 53 wrong) use the first coder on the 53 thin items. The cost analysis uses a blind recode pooling thin and labelled items, 107 answers per policy.

\textbf{Study 3.} 350 planned runs on Gemini 3.1 Pro and GPT-4.1, with GPT-4.1-mini and GPT-5.5 as checks; separately, 40 check runs on Gemini 3.5 Flash and 40 final planned runs on GPT-6.1 Sol and Gemini 3.8 Flash. Study 3's exploratory runs are about 1,100 earlier agent answers from four models.

\textbf{Figure~\ref{fig:summary}.} Study 1 rows are GPT-4.1, Gemini 2.5 Flash and Qwen3.8 27B; Study 2 rows the eleven current models. Study 2 tools lines count agreed answers that stated a figure; the three models that never fetched both plans give 38 of the 44 silent answers on the one-plan line, which is why its interval is wide. The [S] line pools Gemini 3.1 Pro's baseline (fetched the fact 2 of 10) and mention (1 of 10) arms as 3 of 20. Intervals are clustered by question (Study 1) or model (Study 2; the lump-sum line uses Wilson, its cluster interval being degenerate), else Wilson.

\subsection{Coding and statistical methods}
\label{app:stats}
In Study 1 one AI coder (Claude) coded every answer, aided by a model-free text rule that flags a restated thin value with no correcting label nearby. A human recoded a 50-answer sample blind; the full codes matched in 43 of 50 (kappa 0.81). Study 2's two extractors (Claude Opus 5.5, GPT-6.1 Sol) read each answer and record its headline figure, plan, method, mode and any flag or question. A model-based consistency check of 100 agreed answers found no misread headline figure, series, method or mode, and no sign of an extractor favouring its own model. Because both are also tested models, a model outside the panel (gpt-oss-20b) re-extracted all 160 of their open answers: the 23 it marked as not flagged all showed both plans, labelled, and none committed silently. gpt-oss-20b is also the data-match model and the small judge.

Intervals are 95\%: bootstrapped with questions as clusters for Study 1, clustered by model where Study 2 models differ (with 7--11 model clusters these are approximate and likely too narrow), and Wilson for counts of runs. Tests are exact (one-sided Fisher for planned directional contrasts, McNemar or sign), Holm-corrected within Study 3's planned contrasts and Study 2's per-model order tests; unplanned contrasts are marked exploratory. McNemar tests treat answers as independent; question-level sign tests are the conservative reading and are given alongside. Study 3's Fisher tests treat repeated runs of one prompt as independent draws and pool its two cases; they show that an arm changed a model's behaviour on these cases, not that the effect generalises to other cases.

\subsection{Study 1: replications, accurate records and further results}
\label{app:study1}
Table~\ref{tab:repl} gives thin-data behaviour across models. On the 126 answers whose figure was not the asked quantity, the seven models other than GPT-6.1 Sol were wrong in 69--90\% (Gemini 3.8 Flash 99, the small models 106--113 as given); GPT-6.1 Sol was wrong in 30\%.

\begin{table*}[t]
\caption{Thin-data behaviour across models, and with accurate records (fields renamed, entity and plan named). The 42 are 14 questions x 3 models (Study 1) or x 3 samples of one model. The three small models' rows come from one arm-hidden recode of both arms (their first coding differs by at most 3); replications used larger output limits.}
\label{tab:repl}
\small
\begin{tabularx}{\textwidth}{L{1.68}L{0.84}L{0.84}L{0.90}L{1.02}L{0.72}}
\toprule
\textbf{Model (samples per item)} & \textbf{Record} & \textbf{Precise mislabels (of 42)} & \textbf{Wrong, figure not the asked quantity (of 126)} & \textbf{Flagged or asked, same answers (of 126)} & \textbf{Asked (of 162)} \\
\midrule
Study 1: GPT-4.1, Gemini 2.5 Flash, Qwen3.8 27B (1) & As given & 26 & 96 & 18 & 1 \\
GPT-6.1 Sol (3) & As given & 10 & 38 & 88 & 43 \\
Gemini 3.8 Flash (3) & As given & 27 & 99 & 12 & 1 \\
Llama 3.1 8B (3) & As given / accurate & 34 / 21 & 109 / 91 & 12 / 25 & 0 / 0 \\
Mistral Nemo 12B (3) & As given / accurate & 35 / 25 & 113 / 100 & 4 / 20 & 1 / 0 \\
Qwen3 8B (3) & As given / accurate & 32 / 21 & 106 / 92 & 16 / 21 & 3 / 4 \\
\bottomrule
\end{tabularx}
\end{table*}

\textbf{Accurate-record arm.} The arm was planned; departing from the plan, the records also named the entity and plan explicitly. The planned test, a question-level sign test, was not significant for precise mislabels (101 to 67 of 126; 8 questions fewer, 2 more; p = 0.11) and was for wrong answers (328 to 283 of 378; p = 0.039); a Holm correction added after the plan leaves neither significant. The unadjusted question-clustered intervals (-48 to -8 and -23 to -2 points; a percentile bootstrap over 14 questions, which tends to be too narrow with so few clusters) exclude zero, so we read the arm as suggestive, not confirmatory. The model-free rule, which cannot see the arm, moved the same way (mislabel candidates 105 to 43 of 378; sign test p = 0.002). The split by record type is exploratory; on the asset-change example mislabels went from 6 to 9 of 9. On vague questions the removed mislabels became answers that named the figure's true measure without the alternative, which the codebook counts as wrong; coders saw each record, so this shift is partly by definition. Removing the instruction `If data is provided, use it' did not reduce mislabels.

\textbf{Explicit-field restatement.} Wrong answers 131 to 130 of 558 paired answers (90\% interval -1.7 to +1.4 points); on thin data 96 to 94 of 162, where an effect of about 5 points cannot be excluded.

\textbf{Self-chosen context (Study 2, tools).} When models chose what entered their context, the evidence is weaker: answers that fetched one plan series were silent on the plan in 44 of 65 against 4 of 80 when both were fetched, but this pooled contrast mostly reflects which models never fetched both (38 of their 42 answers). Within the four models that fetched one plan in some answers and both in others, silence was 6 of 23 against 1 of 43, five of the six from GLM-5.3. In exploratory agent runs a tool that reported only the lump-sum annualised return misled 35 of 72 SIP answers.

\subsection{Study 2: further results and coding exceptions}
\label{app:study2}
\textbf{Accuracy on specified items.} gpt-oss-120b was right on 58\% with tools; Llama 4 Maverick on 69\% with the table and 13\% with tools.

\textbf{Silence.} The extractors disagreed on some of Claude Opus 5.5's and GPT-6.1 Sol's tool answers; counting every disagreement as silent gives upper bounds of 10\% and 30\%. If Gemini 3.8 Flash's five unworded side-by-side layouts are coded as flagged, the five models give 135 of 144 silent agreed answers. Nine of the 149 fall on Mirae's mode items, where mode is immaterial, so they are not frame collapse.

\textbf{Order.} Kimi K3 followed the first column only partly, as it also favoured Direct. Among the seven models that followed the first column, 49 of 50 pairs flipped (a subset selected on the outcome); a pair flips when the headline follows the first column in both orders. In one model the layout also set a committed answer: Grok 4.7 called Regular ``the standard plan'' in 8 of 10 answers that listed it first and none that listed it second (exploratory).

\textbf{Lump-sum default.} The 81 agreed silent mode answers that assumed a lump sum came from seven models, most from Llama 4 Maverick and Qwen3.5-122B; 72 of 72 on material items; over all answers, including four tool calls printed as text, 83 of 85. Three models favoured the Direct plan whatever the column order (right only when the investor's records say Direct).

\subsection{Study 3: further results}
\label{app:study3}
\textbf{Defaults.} With residency removed, Gemini 3.1 Pro filled a residency field with `resident' in all 15 runs that read the profile. The compute tool that produced the resident-based figure did not, by design, take residency.

\textbf{Gathering arms.} Field-named profile tools without a demand fetched the fact in 9 of 10 runs (correct in 8). With a demand for the decisive fact, Gemini 3.1 Pro answered correctly whether its arguments were named (15 of 15 per case) or uninformatively named (5 of 5 per case). With the fact missing, it filled a required residency field with `resident'.

\textbf{GPT-4.1.} It ignored tool arguments (1 of 10). With the facts preloaded it stayed wrong on the original question (0 of 5). It applied the 30\% withholding only when asked for the amount after tax, in the question itself or in a follow-up turn (10 of 10, five each; one-sided Fisher p = 3.3 $\times$ 10$^{-4}$). Four of those ten were within Rs. 120 of the reference and six omitted the 4\% cess (Rs. 580--725 high, about 0.1\% of the credit, against Rs. 16,484 when the withholding is missed), so under strict scoring, which counts a missing cess as wrong, the contrast is 4 of 10 against 0 of 5 (p = 0.15). Because this rewords the user's question rather than adding a platform control, we read it as evidence that GPT-4.1 had the rate but did not apply it unprompted, not as a deployable fix. When asked directly, GPT-4.1 justified the 30\% rate with a superseded 36-month holding test.

\textbf{Rule version.} The current rule as prompt text gave GPT-4.1 5 of 5 against 0 of 5 with the facts alone (p = 0.004), and a tool computing the tax under it 5 of 5. The facts-alone arm was appended to the plan after 140 earlier runs but before its own, and neither plan specified this Fisher contrast. The plans set threshold readings; a Holm correction over nine Fisher contrasts, added later, keeps six, including both demand contrasts and the rule as text for GPT-4.1, and drops three. In exploratory runs, required plan, method and horizon fields gave no violation in 144 runs.

\textbf{Frontier checks.} GPT-6.1 Sol and Gemini 3.8 Flash were also correct under strict scoring, and in all 20 runs with a demand. Zero failures in 20 baseline runs, 10 per model, gives 95\% Wilson upper limits on the per-run failure probability of 16\% (pooled, treating runs as independent and the two models as sharing one rate) and 28\% per model; it says nothing about other cases.

\subsection{Judges: all arms and tests}
\label{app:judges}
Table~\ref{tab:judgesfull} gives both judges in every arm. The question-only arm was planned together with an arm whose notes sometimes stated the verdict (40 of 44 named for Sol, 28 for gpt-oss-20b); the neutral data and dictionary arms replaced it after a first review, pre-specified before their own runs; the arms without the reading were planned and use the reworded rubric.

\textbf{Tests.} With the reading, dictionary against data alone for Sol: answer-level McNemar p = 1.5 $\times$ 10$^{-6}$; 9 of 13 questions improved, none worsened, sign p = 0.004; Sol rated all 44 mislabels wrong. Data alone against question alone: 17 against 14 of 44 (p = 0.55). Without the reading, dictionary against data alone: p = 7.6 $\times$ 10$^{-5}$; 6 questions improved, none worsened, sign p = 0.03; it passed no wrong precise answer. The small judge passed 43 of 70 wrong vague answers without the reading.

\textbf{Data alone (post hoc).} The data alone can hurt. Shown the data, Sol passed 28 of 70 wrong answers to vague questions against 4 without it (precise: 15 against 14). Over all 96 wrong answers, 7 questions moved towards passing and 3 away (sign p = 0.34); on vague answers alone, 6 and none (sign p = 0.03; post hoc). A simpler check, that every quoted percentage appears in the tool output, passed 31 of 35 misleading SIP answers in exploratory agent runs.

\begin{table*}[t]
\caption{Judges on the 162 Study 1 thin answers. Mislabels named: codebook mislabels the judge labelled as such (the planned outcome); 44 mislabels cover all wordings (26 precise, 18 vague). Wrong rated correct: codebook-wrong answers the judge passed (96 = 26 precise + 70 vague). Acceptable rejected: right, flagged or asked answers called wrong.}
\label{tab:judgesfull}
\small
\begin{tabularx}{\textwidth}{L{0.91}L{0.84}L{1.19}L{1.05}L{0.98}L{0.98}L{1.05}}
\toprule
\textbf{Judge} & \textbf{Told the intended reading} & \textbf{Sees} & \textbf{Mislabels named (of 44)} & \textbf{Wrong rated correct: precise (of 26)} & \textbf{Wrong rated correct: vague (of 70)} & \textbf{Acceptable rejected (of 61)} \\
\midrule
GPT-6.1 Sol & Yes & Question & 14 & 14 & 4 & 15 \\
GPT-6.1 Sol & Yes & + data & 17 & 15 & 28 & 15 \\
GPT-6.1 Sol & Yes & + data + dictionary & 40 & 0 & 0 & 23 \\
GPT-6.1 Sol & No & + data & 23 & 5 & 48 & 15 \\
GPT-6.1 Sol & No & + data + dictionary & 40 & 0 & 42 & 17 \\
gpt-oss-20b & Yes & Question & 5 & 17 & 9 & 37 \\
gpt-oss-20b & Yes & + data & 3 & 16 & 21 & 23 \\
gpt-oss-20b & Yes & + data + dictionary & 24 & 4 & 12 & 21 \\
gpt-oss-20b & No & + data & 7 & 8 & 53 & 11 \\
gpt-oss-20b & No & + data + dictionary & 24 & 3 & 43 & 10 \\
\bottomrule
\end{tabularx}
\end{table*}

\subsection{Data-match check}
\label{app:datamatch}
On labelled data the check left wrong answers at a similar level (8 to 9) while asking in 21 of 53. On the cost score (blind recode; thin and labelled items pooled, 107 answers per policy) it beats both doing nothing and always asking only when an unnecessary question costs between a half and about one wrong answer; at a half it ties always asking (33 against 33) and at one it beats doing nothing by a single point (46 against 47). Always asking would have scored 41 of 53 on appropriate behaviour.

\subsection{Claims, evidence and status}
\label{app:claims}
Table~\ref{tab:claims} lists each main claim with its evidence, scoring, status and scope.

\begin{table*}[t]
\caption{Claims, evidence and status. A manipulation varies one element with everything else fixed; a comparison contrasts models or items.}
\label{tab:claims}
\small
\begin{tabularx}{\textwidth}{L{1.35}L{1.20}L{0.75}L{0.75}L{0.95}}
\toprule
\textbf{Claim} & \textbf{Evidence and design} & \textbf{Scoring} & \textbf{Status} & \textbf{Scope} \\
\midrule
F1. When a lookup tool returns one ambiguously named figure in reply to the question, models attach the user's label to it or silently answer another reading: (i), (ii) & Study 1: thin vs labelled data; accurate-record arm; five replications & Codebook; blind, model-free and human checks & Design fixed in advance; codebook after collection; accurate-record arm planned (entity naming a deviation) & 7 of 8 models (GPT-6.1 Sol mostly flagged or asked; accurate-record arm: 3 small open-weight); 14 questions whose thin figure was not the asked quantity; illustrative figures \\
F2. With plan or mode open, several models answer without naming it (ii) & Study 2: open vs specified items (comparison across 11 models) & Code, two extractors & Planned & Two dimensions, 10 funds \\
F3. Layout sets which plan leads (iii) & Study 2: counterbalanced column order (manipulation) & Code & Planned; committed-answer effect exploratory & 9 of 10 models; committed answers in one \\
F4. Two agents selected for failing omitted conditions they could fetch (iv); a demand for the decisive fact made Gemini 3.1 Pro answer correctly; GPT-4.1 needed the current rule, or an explicit request for the bank credit after TDS & Study 3: one-element arms (manipulation) & Code and tool logs & Pre-specified threshold readings; six of nine Fisher contrasts survive a Holm correction added later & Two cases; models selected for failing; did not transfer to two frontier models (40 runs) \\
F5. A judge named most mislabels when given an accurate data dictionary, with or without the intended reading; without the reading it passed silent readings of vague questions & Judge arms on Study 1 answers (manipulation) & Codebook as reference & Planned outcome; the data arms with the reading added after a first review & Two judges, each with and without the reading (rubric reworded without it); 162 answers \\
F6. Checking data against the question cuts errors, mostly by asking & Data-match check vs baseline (paired) & Codebook; blind recode & Planned; scoring deviated & One open-weight model; always asking scores higher \\
\bottomrule
\end{tabularx}
\end{table*}

\subsection{Controls in full}
\label{app:controls}
Table~\ref{tab:controlsfull} expands Table~\ref{tab:controls}. Where the platform owns the data, the first control is a data contract; a model-side check is the backstop for third-party feeds. The proposed layered design loads the facts the rule takes as inputs (row 1), supplies the rule in force with its legal basis (row 2) and checks the data against the question (row 3).

\begin{table*}[t]
\caption{Controls for a financial assistant: jurisdiction-agnostic control, failure addressed, an Indian example, and the evidence our studies give for it. Rows 1 and 2 have planned evidence for the failing models only; row 4 has planned evidence for the data dictionary, not for reference answers; row 3 has planned evidence only for a prompt-level check that over-asked, and exploratory evidence for metadata; rows 5--8 are untested.}
\label{tab:controlsfull}
\small
\begin{tabularx}{\textwidth}{L{0.20}L{1.35}L{0.60}L{1.35}L{1.50}}
\toprule
\textbf{\#} & \textbf{Control (any jurisdiction)} & \textbf{Failure addressed} & \textbf{Example implementation in India} & \textbf{Evidence here} \\
\midrule
1 & Make each decisive fact a required, typed input taken from the investor's records (or asked of the investor where only they hold it, such as gains booked elsewhere), enforced in code; never accept a model-supplied default & (iv) & Consolidated account statements; tax status on each folio & For Gemini 3.1 Pro a demand for the decisive fact fetched it and gave correct answers (two cases); GPT-4.1 fetched it under a checklist but stayed wrong unless the current rule was supplied or the user explicitly asked for the bank credit after TDS (a diagnostic, not a control); a model filled a required field with `resident' when the fact was missing; frontier models gathered it unasked (20 of 20), so for them the control is a safeguard against regression (untested), not a fix shown to be needed \\
2 & Supply rules from a dated source and compute the tax in code & (iv) & A rulebook re-tested when the Budget, tax notifications or SEBI circulars change & Rule as text 5 of 5 vs 0 of 5 on the version case; a tool with the superseded rule was relayed 5 of 5 (one model); legal-basis text not significant after Holm \\
3 & Give data fields metadata saying what they measure; check them against the question, with a rule that answers when they match (the check tested here lacked one and over-asked) & (i) & Metadata separating total-return from price indices, net flows from asset changes & Accurate records cut mislabels where they named a different entity (exploratory, 3 small models), not for close measures; the data-match check (a prompt, not metadata) cut wrong answers 38 to 9 but asked back 33 times and lost all 11 plainly correct answers \\
4 & Evaluate with the data's meaning (tested); add code-checked reference answers for each reading (untested as a judge input) & Judges passing wrong answers & Pre-release tests of extracted figures against computed values & An accurate, author-written dictionary let a strong judge name 40 of 44 mislabels (i) with or without the intended reading (17 to 23 of 61 false alarms, against 15 with the data alone); without the reading, silent readings (ii) passed (42 of 52); a check that quoted percentages appear in the tool output passed 31 of 35 misleading answers (exploratory) \\
5 & Label every figure with the interpretation used: variant, measure, period, as-of date, fees, tax basis & All four & State plan, option, return method and pre- or post-tax & Not tested (the central principle) \\
6 & Order alternatives by the user's own holding, not by layout or search rank & (iii), (ii) & The investor's plan; else a fixed, disclosed order, never the plan paying the seller more & Not tested; effect observed (Study 2) \\
7 & Keep servicing separate from advice & Advice in servicing & Execution-only platforms route suitability questions to a registered adviser & Not tested; outside the SIP question, 11 of 144 suitability answers recommended outright, 33 conditionally \\
8 & Log the interpretation, facts and rule version of each answer & Audit trail & A per-answer record for the responsibility SEBI places on firms & Not tested \\
\bottomrule
\end{tabularx}
\end{table*}

\subsection{Further limitations}
\label{app:limits}
\textbf{One market.} All three studies use Indian mutual funds, so they cannot show how the failures vary across markets. Models may know some markets' products and rules better than others, for example markets that are better represented in their training data, and less familiar markets may produce more defaults and outdated rules. Repeating the studies in other markets, from widely documented ones such as the US and UK to smaller markets with less public documentation of their products and tax rules, would show how much performance depends on the market. The taxonomy would need further values there, such as real versus nominal returns and tax wrappers.

\textbf{Other limits.} 53 of Qwen3.8 27B's Study 1 answers may have been truncated (none a thin-data wrong answer). Study 3's rule tool cited the repealed Act's section (figures unaffected). Two judges were tested without the intended reading, with a reworded rubric. A planned resolution-layer test on Study 2 was stopped after 6 of 148 items because its pre-step assumed one default plan, whereas the right plan is the one in the investor's records.

\end{document}